\documentclass[conference]{IEEEtran}
\IEEEoverridecommandlockouts

\usepackage{cite}
\usepackage{amsthm}
\usepackage{amsmath,amssymb,amsfonts}
\usepackage{graphicx}
\usepackage{textcomp}
\usepackage{xcolor}
\def\BibTeX{{\rm B\kern-.05em{\sc i\kern-.025em b}\kern-.08em
    T\kern-.1667em\lower.7ex\hbox{E}\kern-.125emX}}

\usepackage{algorithm}
\usepackage{algpseudocode}
\usepackage[utf8]{inputenc} 
\usepackage[T1]{fontenc}    
\usepackage{hyperref}       
\usepackage{url}            
\usepackage{booktabs}       
\usepackage{amsfonts}       
\usepackage{nicefrac}       
\usepackage{microtype}      

\usepackage{adjustbox}
\usepackage{balance}
\usepackage{graphicx}
\usepackage{subfigure}
\usepackage{booktabs} 
\usepackage{amsmath,amssymb,amsfonts}
\usepackage{graphicx}
\usepackage{textcomp}
\usepackage{xcolor}
\usepackage{booktabs}
\usepackage{multirow}
\usepackage{cleveref}
\usepackage{times}
\usepackage{fancyhdr}

\usepackage{caption}
\usepackage{tabularx}
\usepackage{xcolor}
\usepackage{tablefootnote}
\usepackage{ragged2e}

\def\reals{\mathbb{R}}

\def\comp{\raise 1pt \hbox{$\scriptstyle\circ$}}

\def\argmin{\mathop{\rm argmin}\limits}

\def\minimize{\mathop{\rm min}\limits}

\def\st{\mathop{\rm subject\ to}}

\def\upto{{\raise 1pt \hbox{$\scriptstyle \,\nearrow\,$}}}
\def\downto{{\raise 1pt \hbox{$\scriptstyle \,\searrow\,$}}}

\begin{document}

\newtheorem{theorem}{Theorem}
\newtheorem{definition}{Definition}

\title{Decomposition of Difficulties in Complex Optimization Problems Using a Bilevel Approach}

\author{\IEEEauthorblockN{Ankur Sinha}
\IEEEauthorblockA{
\textit{Centre for Data Science and AI}\\
Indian Institute of Management\\
Ahmedabad, Gujarat, India 380015 \\
asinha@iima.ac.in}
\and
\IEEEauthorblockN{Dhaval Pujara}
\IEEEauthorblockA{
\textit{Centre for Data Science and AI}\\
Indian Institute of Management\\
 Ahmedabad, Gujarat, India 380015 \\
dhavalp@iima.ac.in}
\and
\IEEEauthorblockN{Hemant Kumar Singh}
\IEEEauthorblockA{ 
\textit{School of Engineering and Technology}\\
The University of New South Wales\\
Canberra, ACT 2610, Australia\\
h.singh@unsw.edu.au}
}


\maketitle

\IEEEpubidadjcol

\begin{abstract}
Practical optimization problems may contain different kinds of difficulties that are often not tractable if one relies on a particular optimization method. Different optimization approaches offer different strengths that are good at tackling one or more difficulty in an optimization problem. For instance, evolutionary algorithms have a niche in handling complexities like discontinuity, non-differentiability, discreteness and non-convexity. However, evolutionary algorithms may get computationally expensive for mathematically well behaved problems with large number of variables for which classical mathematical programming approaches are better suited. In this paper, we demonstrate a decomposition strategy that allows us to synergistically apply two complementary approaches at the same time on a complex optimization problem. Evolutionary algorithms are useful in this context as their flexibility makes pairing with other solution approaches easy. The decomposition idea is a special case of bilevel optimization that separates the difficulties into two levels and assigns different approaches at each level that is better equipped at handling them. We demonstrate the benefits of the proposed decomposition idea on a wide range of test problems. 
\end{abstract}

\vspace{2mm}


\section{Introduction}
Diverse scientific and engineering disciplines grapple with a multitude of intricate optimization challenges. The complexity inherent in such optimization problems can manifest through various characteristics of the feasible region, including but not limited to, high dimensionality, discontinuity, discreteness, and nonconvexity. A comprehensive body of literature addressing optimization problems offers an array of methodologies to tackle these complexities. These methodologies primarily fall into two categories: the classical mathematical programming approach and metaheuristic approaches. The classical mathematical programming approaches utilize the known properties of a problem in the search, and typically provide some guarantees regarding optimality of the solutions achieved. They
encompasses a broad range of optimization methods within the realm of linear programming, mixed-integer programming and non-linear programming \cite{luenberger1984linear}. 

Conversely, the metaheuristic approaches typically do not provide any guarantees regarding the optimality of the solutions achieved. However, they also do not require assumptions on the underlying functions, and hence applicable to a much wider range of practical problems. This capability also comes with a cost that metaheuristic approaches normally require large numbers of evaluations to reach a competitive solution. Metaheuristic themselves span a range of algorithms such as evolutionary algorithms~(EAs), ant colony optimization, simulated annealing, and particle swarm optimization that have shown potential in handling certain classes of optimization problems in which classical approaches suffer \cite{gendreau2010handbook}. Among these, EAs are the most widely used, and hence we'll focus on them as a representative metaheuristic for the remainder of the paper. 

\begin{figure*}[hbt]
\centering
\includegraphics[width=0.80\textwidth]{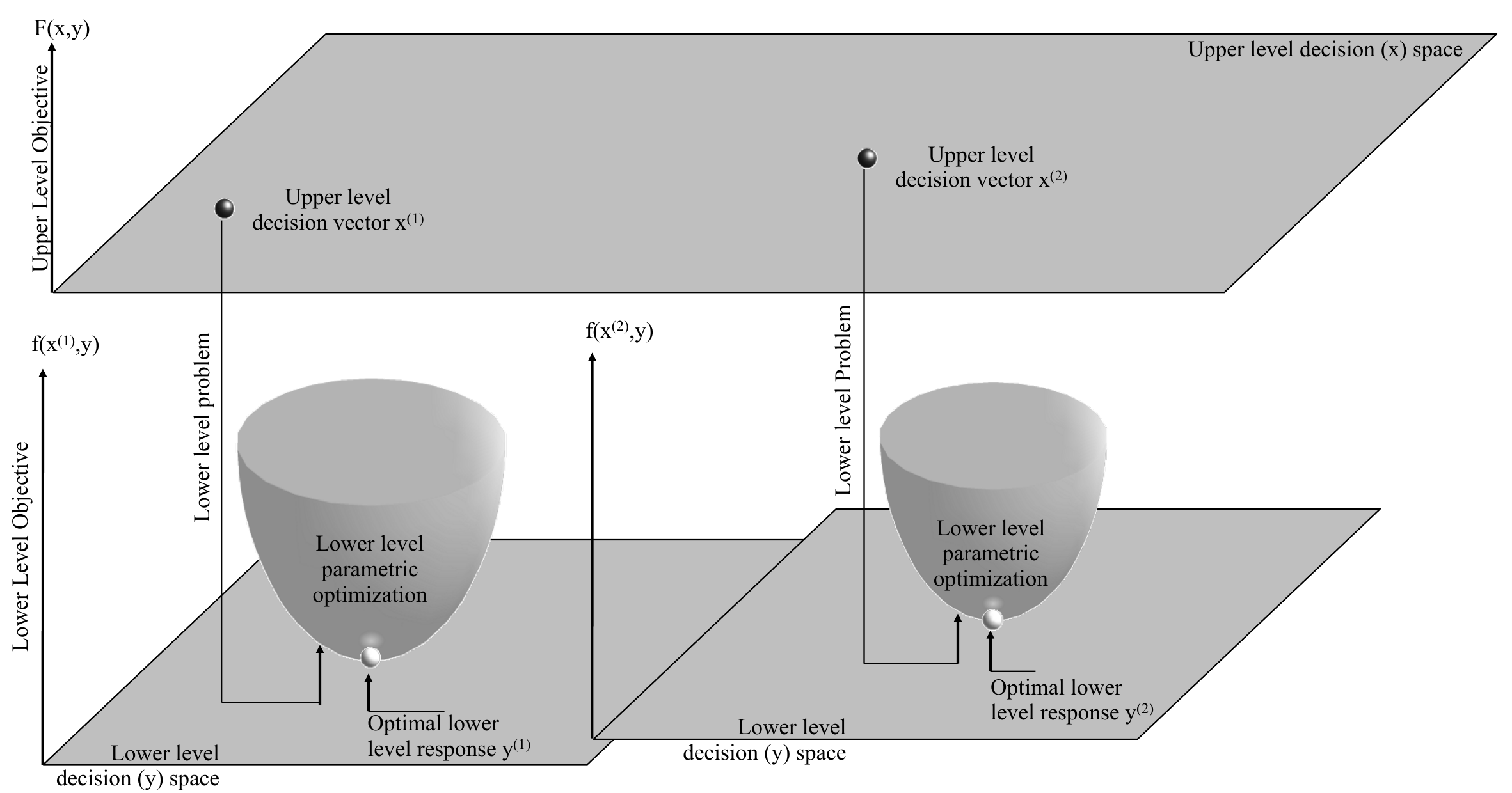}
\caption{A sketch of a bilevel problem showing the linkage between the upper and the lower levels.}
\label{fig:sketch}
\end{figure*}

While both mathematical programming and evolutionary approaches can individually address certain types of complexities, their efficacy is contingent on the nature of the optimization problem. For instance, well-structured problems adhering to regularities like continuity, differentiability, or convexity can be effectively solved using classical mathematical methods, even when posed with a substantial number of variables and constraints~\cite{eiselt2007linear}. However, these methods prove less effective when confronted with non-regularities such as discontinuity, non-differentiability, or non-convexity, where EAs yield superior results, albeit for problems with lower dimensionality \cite{brownlee2011clever}. Real-world optimization problems frequently exhibit an unstructured nature, encompassing both difficulties, i.e. non-regularities and high dimensionality.

In response to such complex scenarios, this study advocates a decomposition strategy wherein difficulties are bifurcated into two levels and addressed separately using an appropriate approach. To implement this strategy, our study tackles a general optimization problem within the framework of a specialized class of optimization known as bilevel optimization \cite{my-ieeetec18,dempe02}. We propose a Bilevel Optimization-based Decomposition~(BOBD) method designed to solve standard optimization problems (single objective and single level) with a variety of complexities. We force a bilevel decomposition of the standard optimization problem and break it in two levels. Thereafter, we utilize both an EA as well as a classical approach to solve this problem. Interestingly, using this decomposition strategy, we are able to solve problems that are otherwise difficult to solve if attempted using EAs or classical approaches alone. Solving the problem in a bilevel framework also allows us to exploit two important mappings, i.e., the lower level reaction set mapping and the lower level optimal value function mapping to reduce the computational requirements.

We assess the performance of BOBD on a test suite comprising 10 problems. Subsequently, we compare the convergence properties of the suggested BOBD approach with the classical mathematical programming approach and the evolutionary approach. This comparison is based on an empirical analysis of the results obtained from solving the test suite problems using methods corresponding to each approach.

The remainder of the paper is structured as follows: Section II outlines the proposed decomposition strategy, including a description of bilevel optimization. An illustrative example is also provided in this section to explain the idea behind BOBD. Section III discusses the description and development of 10 test problems with varying levels of difficulties. Section IV demonstrates the procedure of BOBD and discussed the results obtained by employing multiple optimization approaches to solve the test  problems. Section V presents concluding remarks and future work.

\section{Bilevel Optimization}
Bilevel optimization is a specialized class of optimization problems characterized by a hierarchical structure comprising two distinct optimization levels: an upper-level optimization task with a lower-level optimization task embedded within the upper level as a constraint \cite{BrMc73,Ruuska2011ConnectionsBS,my-ieeetec18}. The upper-level problem is commonly denoted as the leader's optimization problem, while the lower-level problem is regarded as the follower's optimization problem. Within the bilevel optimization framework, both the leader and the follower possess their individual objectives and constraints. The leader, possessed with comprehensive knowledge of the follower's expected response, initiates the process, to which the follower reacts optimally. Consequently, for any given leader action, the corresponding optimal follower response serves as a feasible solution for the bilevel problem provided that the leader's constraints are satisfied. The primary objective is to select the optimal action plan (solution) for the leader.

A schematic of bilevel optimization is shown in Figure \ref{fig:sketch}, wherein the potential set of moves from the leader and follower is represented by the upper-level and lower-level decision spaces, respectively. The leader's action is symbolized by the upper-level decision vector $x$, while the follower's response is indicated by the lower-level decision vector $y$. For a specific $x$, the corresponding lower-level optimization problem constitutes a parametric optimization challenge to be solved with respect to lower-level variables $y$, while $x$ serves as a parameter. For a given $x^o$ if the resulting optimal lower-level decision vector is denoted as $y^o$, then the pair $(x^o, y^o)$ constitutes a feasible bilevel solution for the upper-level optimization problem, provided it adheres to the other constraints inherent in the problem. With this foundational premise, a formal definition of the bilevel optimization problem is given as follows:

\begin{definition}\label{def:bilevel}
For the upper-level objective function $F:\reals^n\times\reals^m \to\reals$ and the lower-level objective function $f:\reals^n\times\reals^m \to\reals$, the bilevel problem is given by 
\begin{align}
\minimize_{x,y} \quad & F(x,y) \label{eq:startOrig}\\
\st &\notag\\  & \hspace{-12mm}y\in \argmin_{y} 
	\lbrace
		f(x,y) : g_k(x,y)\leq 0, k=1,\dots,K, \notag\\ & \hspace{20mm} h_l(x,y) = 0, l=1,\dots,L
	\rbrace\\
 & \hspace{-12mm}G_i(x,y)\leq 0, \quad j=1,\dots,I\\
  & \hspace{-12mm}H_j(x,y) = 0, \quad j=1,\dots,J \label{eq:endOrig}
\end{align}
where $g_k:\reals^n\times\reals^m \to \reals$ and $h_l:\reals^n\times\reals^m \to \reals$ denote the lower level constraints, and  $G_i:\reals^n\times\reals^m \to \reals$ and $H_j:\reals^n\times\reals^m \to \reals$ denote the upper level constraints.
Variables $x$ and $y$ in the above definition may be real or integers as defined in the problem.
\end{definition}

\subsection{Separation of Difficulties in Single Level Optimization Problem }

Single level optimization problems can be considered as a special case of bilevel optimization problems \cite{BrMc73}. In this section, we discuss the procedure of separating the difficulties in single level problem by applying the proposed Bilevel Optimization-based Decomposition (BOBD) method. 

\begin{definition}\label{def:singleLevel}
We represent a general optimization problem that we want to decompose in the form of bilevel optimization problem to exploit the capabilities of two algorithms simultaneously.
\begin{align}
\minimize_{x} \quad & F(x) \label{eq:startSingle}\\
\st &\notag\\
 & \hspace{-12mm}G_i(x)\leq 0, \quad i=1,\dots,I\\
  & \hspace{-12mm}H_j(x) = 0, \quad j=1,\dots,J \label{eq:endSingle}
\end{align}
where $x=(x_1,\ldots,x_n)$ is a solution vector.
\end{definition}

In Definition 2, the single level optimization problem is characterized by $n$ decision variables $(x_1,\ldots,x_n)$. The complexity of this optimization problem increases in the presence of certain non-regularities, like non-linearity, non-convexity, discontinuity, discreteness, non-differentiability, etc. Additionally, high dimensionality, i.e., presence of a large number of decision variables, can contribute to the complexity of the optimization problem as well. Classical mathematical programming and evolutionary methods are inadequate for handling both non-regularities and high dimensionality simultaneously. For regular problems, classical methods handle high dimensionality well, but they often fail for non-regular problems of even small dimensions. On the other hand, evolutionary algorithms handle non-regularity well for problems of low dimensions, but do not scale well for problems with higher dimensions. Interestingly, both approaches have their own strengths, and the proposed bilevel optimization-based decomposition method allows us to solve optimization problems with multiple difficulties by separating them in two levels and then deploying multiple algorithms synergistically. The single level optimization problem can be decomposed as follows. 

Assume that in the given single level optimization problem \eqref{eq:startSingle}-\eqref{eq:endSingle} a chosen set of decision variable $x_{k^+}$ introduces complexity of one type, while the remaining set of variables $x_{k^-}$ introduces complexity of another type. Similarly, assume that the constraints $G_i$ and $H_j$ also introduce complexities of two kinds based on which they can be divided into $G_{i^+}$, $H_{j^+}$ and $G_{i^-}$, $H_{j^-}$, respectively. It is possible to decompose the problem into two levels based on these characteristics, so that the difficulties are separated in two levels.
\begin{align}
\minimize_{x_{k^+},x_{k^-}} \quad & F(x_{k^+},x_{k^-}) \label{eq:startDecom}\\
\st &\notag\\  & \hspace{-12mm} x_{k^-} \in \argmin_{x_{k^-}} 
	\lbrace
		F(x_{k^+},x_{k^-}) : G_{i^-}(x_{k^+},x_{k^-})\leq 0, \notag \\
        & \hspace{32mm} H_{j^-}(x_{k^+},x_{k^-})= 0
	\rbrace\\
 & \hspace{-12mm}G_{i^+}(x_{k^+},x_{k^-})\leq 0\\
  & \hspace{-12mm}H_{j^+}(x_{k^+},x_{k^-})= 0\label{eq:endDecom}
\end{align}
Note that the above formulation would lead to the same optimal solution as the formulation \eqref{eq:startSingle}-\eqref{eq:endSingle}, since the objective functions at both levels have been kept the same.

In the context of bilevel optimization, the leader initiates the process by determining the values of upper-level variables followed by the follower optimally determining its response. In the above scenario, for any given value of $x_{k+}$ a lower level optimization problem can be solved with respect to $x_{k-}$, while keeping $x_{k+}$ fixed. In case the pair $(x_{k^+},x_{k^-})$ satisfies all the constraints in the system it is a feasible solution to the above bilevel formulation and also the single level formulation \eqref{eq:startSingle}-\eqref{eq:endSingle}. Additionally, the bilevel optimal solution coincides with the optimal solution of the single level optimization problem. 

For the above structure, it is possible to deploy an EA for search at the upper level, while a classical algorithm can be applied for search at the lower level. While the EA performs an intensive search based on the evolutionary sampling approach, one can solve the lower level optimization problem for each sample repeatedly to converge to the optimal bilevel solution. This is equivalen to nested approaches that are often deployed in bilevel optimization \cite{mathieu,li2015genetic,my-caor14,angelo13,singh2019nested}. Using an intelligent bilevel approach based on efficient sampling and approximation of mappings \cite{my-ejor17,angelo2014differential,my-joh20,sinha2021solving} in bilevel one can solve such a bilevel problem much more efficiently than a nested approach. Other potential approaches to reduce the computational expense involve hybridization~\cite{islam2017enhanced}, knowledge transfer~\cite{wang2022investigating,chen2021transfer} and surrogate-assisted search~\cite{islam2017surrogate,angelo2019performance,islam2018efficient}. 

In this study, we will demonstrate that for certain classes of optimization problems with multiple difficulties, a synergistic approach based on BOBD leads to a faster and better solution than what can be obtained by using one of the algorithms. Interestingly, this approach has helped us find better solutions for certain single level optimization problems than what has been reported in the literature. Appropriate separation of difficulties and the utilization of suitable methods for each level in the bilevel setup contributes to the effectiveness of the overall solution strategy.

\subsection{Proof-of-concept}
We employ a numerical example to discuss the process of delineating complexities using our approach. Specifically, we examine a single level optimization problem described below:

\begin{align}
\minimize_{x} \quad & F(x)=x_1^{0.6}+x_2^{0.6}+x_3^{0.4}-4 x_3+2 x_4+5 x_5-x_6 \nonumber\\
\st &\notag\\
& x_2-3 x_1-3 x_4=0\nonumber\\
& x_3-2 x_2-2 x_5=0\nonumber\\
& 4 x_4-x_6=0\nonumber\\
& x_1+2 x_4 \nonumber\leq 4\\
& x_2+x_5 \nonumber\leq 4\\
& x_3+x_6 \nonumber\leq 6\\
& x_1 \leq 3, x_3 \leq 4, x_5 \leq 2\nonumber\\
& x_1, x_2, x_3, x_4, x_5, x_6 \geq 0 \nonumber
\end{align}

We separate the difficulties in the above problem as follows: the problem entails a nonconvex objective function under the constraints of three linear inequalities and three linear equalities. The bound constraints give rise to additional inequality constraints. 

Considering each of the six variables and their associated terms in objective function and constraints, it can be observed that variables $x_1$, $x_2$, and $x_3$ cause non-convexity in the objective function. As per the variable classification scheme described in the previous section, we tag these variables as upper level variables: $u = (x_1,x_2,x_3)$. Further, the variables $x_4$, $x_5$, and $x_6$ introduce linear terms in objective function and constraints. Therefore, we tag these variables as lower level variables: $v = (x_4,x_5,x_6)$. Next, we get the following structure after substituting all variables into bilevel optimization structure:
\begin{align*}
\minimize_{u,v} \quad & u_1^{0.6}+u_2^{0.6}+u_3^{0.4}-4 u_3+2 v_1+5 v_2-v_3 \\
\st &\notag\\
v \in \argmin&\left\{\begin{array}{l}
u_1^{0.6}+u_2^{0.6}+u_3^{0.4}-4 u_3+2 v_1+5 v_2-v_3\\
\st \notag\\
u_2-3 u_1-3 v_1=0, \quad u_1+2 v_1 \leq 4\\
u_3-2 u_2-2 v_2=0, \quad u_2+v_2 \leq 4\\
4 v_1-v_3=0, \quad\quad\quad\hspace{3mm} u_3+v_3 \leq 6\\
v_2 \leq 2, \quad\quad\quad\quad\quad\hspace{5.5mm} v_1, v_2, v_3 \geq 0
\end{array}\right\}\\
& \hspace{-16mm} u_1 \leq 3, u_3 \leq 4\\
& \hspace{-16mm} u_1, u_2, u_3 \geq 0
\end{align*}
The difficulties have been separated such that the upper level has a non-convex objective function with only box constraints, while the lower level is a linear program. The above problem can be efficiently handled by an EA at the upper level and a linear programming approach at the lower level. However, as a single level problem it is not easy to solve this simple 6-variable optimization problem either with an EA or a classical non-linear optimization approach. Interestingly, when we solve the optimization problem using the BOBD method we are able to obtain a better solution ($-13.417$) than what is reported ($-11.96$) in the literature \cite{stephanopoulos1975use}. The solution obtained by BOBD ($-13.417$) corresponds to the decision vector $x^{*} = (0.1622, 1.9951, 4, 0.4998, 0.0002, 2.0088)$.


\begin{table*}
\vspace{-5mm}
\caption{Description of the Test Problems (TP1-TP5)}\label{tab1}%
\[
\begin{array}{c l c}
\hline
\rule{0pt}{12pt} 
\multirow{-1.5}{*}{ Problem } &
\multicolumn{1}{c}{\multirow{-1.5}{*}{Formulation}}
& \multirow{-1.5}{*}{Source} \\
\hline
\rule{0pt}{12pt} 
\multirow{-1.5}{*}{ TP1 } &
\multicolumn{1}{c}{\multirow{-1.5}{*}{Presented in Section II }}
& \multirow{-1.5}{*}{\cite{stephanopoulos1975use}} \\
\hline
\addlinespace
\multirow{-11}{*}{ TP2 } & \begin{aligned}
\operatorname{Min} H(x) \hspace{1mm} = \hspace{1mm} &
x_1 \\
\text { s.t. } & 35 x_2^{0.6}+35 x_3^{0.6}-x_1 \leq 0 ; \\
& \hspace{-4mm} -300 x_3+7500 x_5-7500 x_6-25 x_4 x_5+25 x_4 x_6+x_3 x_4=0 ; \\
& 100 x_2+155.365 x_4+2500 x_7-x_2 x_4-25 x_4 x_7-15536.5=0 ; \\
& \hspace{-4mm} -x_5+\ln \left(-x_4+900\right)=0 ; \\
& \hspace{-4mm} -x_6+\ln \left(x_4+300\right)=0 ; \\
& \hspace{-4mm} -x_7+\ln \left(-2 x_4+700\right)=0 ; \\
& 0 \leq x_1 \leq 1000 ;\quad 0 \leq x_2, x_3 \leq 40 ;\quad 100 \leq x_4 \leq 300 ; \quad 6.3 \leq x_5 \leq 6.7 ;\quad 5.9 \leq x_6 \leq 6.4 ;\quad 4.5 \leq x_7 \leq 6.25\\
\end{aligned} &
\multirow{-11}{*}{\cite{Liang2006ProblemDA}}\\
\addlinespace
\hline
\addlinespace
\multirow{-10}{*}{ TP3 } & \begin{aligned}
\operatorname{Min} H(x) \hspace{1mm} = \hspace{1mm} & x_1^{0.6}+x_2^{0.6}+x_3^{0.4}-4 x_3+2 x_4+5 x_5-x_6+\frac{x_3^2}{16}-2 \cos \left(2 \pi x_2\right) \\
\text { s.t. } & x_2-3 x_1-3 x_4=0; \\
& x_3-2 x_2-2 x_5=0; \\
& 4 x_4-x_6=0; \\
& x_1+2 x_4 \leq 4; \\
& x_2+x_5 \leq 4; \\
& x_3+x_6 \leq 6;\\
& x_1 \leq 3 ;\quad x_5 \leq 2 ;\quad x_3 \leq 4; \quad x_1, x_2, x_3, x_4, x_5, x_6 \geq 0\\
\end{aligned} &
\multirow{-10}{*}{\begin{tabular}{c} TP1 \\ Extension \\ \end{tabular}}\\
\addlinespace
\hline
\multirow{-12}{*}{ TP4 } & \begin{aligned}
\operatorname{Min} H(x) \hspace{1mm} = \hspace{1mm} & x_1^{0.6}+x_2^{0.6}+x_3^{0.4}-4 x_3+2 x_4+5 x_5-x_6+\frac{x_3^2}{16}-\frac{x_2^2}{16}-2 \cos \left(2 \pi x_3\right)-2 \cos \left(2 \pi x_2\right)+\sum_{p=1}^P y_p \cdot x_1^{0.6} \\
\text { s.t. } & x_2-3 x_1-3 x_4=0; \\
& x_3-2 x_2-2 x_5=0; \\
& 4 x_4-x_6=0; \\
& x_1+2 x_4 \leq 4; \\
& x_2+x_5 \leq 4; \\
& x_3+x_6 \leq 6 ; \\
& \sqrt{\left(x_1+x_2+x_3\right)}-y_p \leq 0, \hspace{1.5mm} \forall p ; \\
& x_1 \leq 3 ;\quad x_5 \leq 2 ;\quad x_3 \leq 4; \quad 1 \leq y_p \leq 5, \hspace{0.5mm} \forall p; \quad x_1, x_2, x_3, x_4, x_5, x_6 \geq 0 \\
\end{aligned} &
\multirow{-12}{*}{\begin{tabular}{c} TP3 \\ Extension \\ \end{tabular}}\\
\addlinespace
\hline
\addlinespace
\multirow{-13.5}{*}{ TP5 } & \begin{aligned}
\operatorname{Min} H(x) \hspace{1mm} = \hspace{1mm} & x_1-50 \cos \left(2 \pi x_4\right)+\sum_{p=1}^P \frac{y_p{ }^2}{x_4} \\
& \hspace{-7mm}\text { s.t. } 35 x_2^{0.6}+35 x_3^{0.6}-x_1 \leq 0 \text {; } \\
& \hspace{-4mm} -300 x_3+7500 x_5-7500 x_6-25 x_4 x_5+25 x_4 x_6+x_3 x_4=0 ; \\
& 100 x_2+155.365 x_4+2500 x_7-x_2 x_4-25 x_4 x_7-15536.5=0 \text {; } \\
& \hspace{-4mm} -x_5+\ln \left(-x_4+900\right)=0 \text {; } \\
& \hspace{-4mm} -x_6+\ln \left(x_4+300\right)=0 \text {; } \\
& \hspace{-4mm} -x_7+\ln \left(-2 x_4+700\right)=0 \text {; } \\
& x_4^{0.2}+x_5+x_6-y_p \leq 0, \hspace{1.5mm}\forall p ; \\
& 0 \leq x_1 \leq 1000 ;\quad 0 \leq x_2, x_3 \leq 40 ;\quad 100 \leq x_4 \leq 300 ; \\
& 6.3 \leq x_5 \leq 6.7 ;\quad 5.9 \leq x_6 \leq 6.4 ;\quad 4.5 \leq x_7 \leq 6.25 ;\quad 10 \leq y_p \leq 30, \hspace{0.5mm} \forall p \\
\end{aligned} &
\multirow{-13.5}{*}{\begin{tabular}{c} TP2 \\ Extension \\ \end{tabular}}\\
\addlinespace
\hline
\end{array}
\]
\end{table*}


\begin{table*}
\caption{Description of Test Problems (TP6-TP9)}\label{tab2}%
\[
\begin{array}{c l c}
\hline
\rule{0pt}{8pt} 
\multirow{-1}{*}{ Prob. } &
\multicolumn{1}{c}{\multirow{-1}{*}{Formulation}}
& \multirow{-1}{*}{Source} \\

\hline
\addlinespace
\multirow{-14}{*}{ TP6 } &  \begin{aligned}
\operatorname{Min} H(x) =  &
-25\left(x_1-2\right)^2-\left(x_2-2\right)^2-\left(x_3-1\right)^2-\left(x_4-4\right)^2-\left(x_5-1\right)^2-\left(x_6-4\right)^2+\sum_{p=1}^P\left(x_3-y_p\right)^2-\sum_{\mathrm{q}=1}^Q\left(x_5-z_q\right)^2 \\
\text { s.t. }- & \left(x_3-3\right)^2-x_4+4 \leq 0 ; \\
- & \left(x_5-3\right)^2-x_6+4 \leq 0 ; \\
- & x_1-x_2+2 \leq 0 ; \\
& x_1-3 x_2 \leq 2 ; \\
& x_2-x_1 \leq 2 ; \\
& x_1+x_2 \leq 6 ; \\
& y_p-x_3+1 \leq 0,\hspace{1.5mm} \forall p ; \\
& z_q^2-x_3^2-x_5^2 \leq 0,\hspace{1.5mm} \forall q ; \\
& 0 \leq x_1; \quad 0 \leq x_2 ;\quad 1 \leq x_3 \leq 5 ;\quad 0 \leq x_4 \leq 6 ; \quad 1 \leq x_5 \leq 5 ;\quad 0 \leq x_6 \leq 10 ;\quad 0 \leq y_p \leq 5, \hspace{0.5mm} \forall p;\quad 0 \leq z_q \leq 5, \hspace{0.5mm} \forall q\\
\end{aligned} &
\multirow{-14}{*}{\begin{tabular}{c} \text{\cite{floudas1990collection}} \\ Extension \\ \end{tabular}}\\
\addlinespace

\hline
\addlinespace
\multirow{-12}{*}{ TP7 } & \fontsize{6.5}{8}\selectfont \begin{aligned}
\operatorname{Min} H(x)= &
6 x_1+16 x_2-9 x_5+10\left(x_6+x_7\right)-15 x_8+x_9^2+50 \cos \left(\pi x_9\right)-25 \cos \left(\pi x_8\right)-\ln \left(x_8-x_9\right)-\sum_{p=1}^P\left(y_p-x_9\right)^2+\sum_{q=1}^Q\left(z_q-x_8\right)^2 \\
\text { s.t. } & x_1+x_2-x_3-x_4=0 ; \\
& x_3+x_6-x_5=0 ; \\
& x_4+x_7-x_8=0 ; \\
& 0.03 x_1+0.01 x_2-x_3 x_9-x_4 x_9=0 ; \\
& x_3 x_9+0.02 x_6-0.025 x_5 \leq 0 ; \\
& x_4 x_9+0.02 x_7-0.015 x_8 \leq 0 ; \\
& x_9^2-y_p^2 \leq 0,\hspace{1.5mm} \forall p ; \\
& x_8^2-z_q^2 \leq 0,\hspace{1.5mm} \forall q ; \\
& 0 \leq x_1, x_2, x_6 \leq 300 ;\quad 0 \leq x_3, x_5, x_7 \leq 100 ;\quad 0 \leq x_4, x_8 \leq 200; \quad 0.01 \leq x_9 \leq 0.03 ;\quad 0 \leq y_p \leq 1, \hspace{0.5mm} \forall p;\quad 1 \leq z_q \leq 200, \hspace{0.5mm} \forall q\\
\end{aligned} &
\multirow{-12}{*}{\begin{tabular}{c} \text {\cite{Liang2006ProblemDA}} \\ Extension \\ \end{tabular}}\\
\addlinespace

\hline
\addlinespace
\multirow{-18}{*}{ TP8 } & \begin{aligned}
\operatorname{Min} H(x) \hspace{1mm} = \hspace{1mm} & 5 \sum_{i=1}^4 x_i-5 \sum_{i=1}^4 x_i^2-\sum_{i=5}^{13} x_i-20 e^{-0.1 \sqrt{\sum_{i=1}^4 x_i^2}}-e^{0.25 \sum_{\mathrm{i}=1}^4 \cos \left(2 \pi x_i\right)}+\sum_{p=1}^P\left(y_p^2+\sum_{\mathrm{i}=1}^4 \cos \left(2 \pi x_i\right)\right) \\
\text { s.t. } \quad & 2 x_1+2 x_2+x_{10}+x_{11} \leq 10; \\
& 2 x_1+2 x_3+x_{10}+x_{12} \leq 10 ; \\
& 2 x_2+2 x_3+x_{11}+x_{12} \leq 10 ; \\
& x_{10}-8 x_1 \leq 0 ; \\
& x_{11}-8 x_2 \leq 0 ; \\
& x_{12}-8 x_3 \leq 0 ; \\
& x_{10}-x_5-2 x_4 \leq 0 ; \\
& x_{11}-x_7-2 x_6 \leq 0 ; \\
& x_{12}-x_9-2 x_8 \leq 0 ; \\
& \sum_{\mathrm{i}=1}^4 \cos \left(2 \pi x_i\right)-y_p \leq 0, \hspace{1mm} \forall p ; \\
& 0 \leq x_i \leq 3 \hspace{1.5mm}(i=1, \ldots, 4); \quad 0 \leq x_i \leq 1 \hspace{1.5mm}(i=5, \ldots, 9) ;\quad 0 \leq x_i \leq 100 \hspace{1.5mm}(i=10,11,12);\\
& 0 \leq x_{13} \leq 1 ; \quad -5 \leq y_p \leq 5, \hspace{0.5mm} \forall p
\end{aligned} &
\multirow{-18}{*}{\begin{tabular}{c} \text {\cite{Liang2006ProblemDA}} \\ Extension \\ \end{tabular}}\\
\addlinespace
\hline

\addlinespace
\multirow{-13}{*}{ TP9 } & \begin{aligned}
\operatorname{Min} H(x) \hspace{1mm} = \hspace{1mm} & x_1+x_2+x_3+\sum_{p=1}^P\left(\tan \left(y_p\right)-15 \cos 2 \pi\left(x_1+x_2+x_3\right)\right)^2 \\
\text { s.t. } & 0.0025 x_4+0.0025 x_6 \leq 1 ; \\
& 0.0025 x_5+0.0025 x_7-0.0025 x_4 \leq 1 ; \\
& 0.01 x_8-0.01 x_5 \leq 1 ; \\
& \hspace{-4mm}-x_1 x_6+100 x_1+833.33 x_4 \leq 83333.33; \\
& \hspace{-4mm}-x_2 x_7+1250 x_5-1250 x_4+x_2 x_4 \leq 0; \\
& \hspace{-4mm}-x_3 x_8-2500 x_5+x_3 x_5+1250000 \leq 0; \\
& \hspace{-1mm}\tan \left(y_p\right)-\ln \left(x_1+x_2+x_3\right) \leq 0 \quad \forall p;\\
& \hspace{-4mm}-\tan \left(y_p\right)-\ln \left(x_1+x_2+x_3\right) \leq 0 \quad \forall p; \\
& 100 \leq x_1 \leq 10000; \quad 1000 \leq x_i \leq 10000 \hspace{1.5mm} (i=2,3); \quad 10 \leq x_i \leq 1000  \hspace{1.5mm}(i=4, \ldots, 8); \quad -\pi / 2 \leq  y_p \leq \pi / 2, \hspace{0.5mm} \forall p\\
\end{aligned} &
\multirow{-13}{*}{\begin{tabular}{c} \text {\cite{Liang2006ProblemDA}} \\ Extension \\ \end{tabular}}\\
\addlinespace
\hline
\end{array}
\]
\end{table*}

\begin{table*}
\centering
\caption{Description of Test Problems (TP10)}\label{tab3}%
\[
\begin{array}{@{}c l c@{}}
\hline
\rule{0pt}{8pt} 
\multirow{-1}{*}{ Prob. } &
\multicolumn{1}{c}{\multirow{-1}{*}{Formulation}}
& \multirow{-1}{*}{Source} \\

\hline
\addlinespace
\multirow{-15}{*}{ TP10 } &  \begin{aligned}
\operatorname{Min} H(x) =  &
37.293239x_1 + 0.8356891x_1x_5 + 5.3578547x_3^2 - 40792.14 + \sum_{p=1}^{P}(y_p - x_1 - x_3)^2- 150\sum_{q=1}^{Q}\cos(2\pi z_q) \quad \quad \quad \quad \quad \\
\text{s.t.}\quad & 0.0056858x_2x_5 - 0.0022053x_3x_5 + 0.0006262x_1x_4 \le 6.665593; \\
&  \hspace{-4mm} -0.0056858x_2x_5 + 0.0022053x_3x_5 - 0.0006262x_1x_4 \le 85.334407; \\
&  0.0071317x_2x_5 + 0.0021813x_3^2 + 0.0029955x_1x_2 \le 29.48751; \\
& \hspace{-4mm} -0.0071317x_2x_5 - 0.0021813x_3^2 - 0.0029955x_1x_2 + 9.48751 \le 0; \\
& 0.0047026x_3x_5 + 0.0019085x_3x_4 + 0.0012547x_1x_3 \le 15.699039; \\
& \hspace{-4mm} -0.0047026x_3x_5 - 0.0019085x_3x_4 - 0.0012547x_1x_3 + 10.699039 \le 0; \\
& y_p - \ln(x_1 + x_3 + 1) \le 0, \hspace{1mm} \forall p; \\
& z_q^3 - x_1^3 - x_3^3 - x_5^3 \le 0, \hspace{1mm} \forall q; \\
& 78 \le x_1 \le 102; \quad 33 \le x_2 \le 45; \quad 27 \le x_3, x_4, x_5 \le 45; \quad 0 \le y_p \le 5, \hspace{0.5mm} \forall p; \quad -5 \le z_q \le 5, \hspace{0.5mm} \forall q
\end{aligned} &
\multirow{-15}{*}{\begin{tabular}{c} \text {\cite{Liang2006ProblemDA}} \\ Extension \\ \end{tabular}}\\
\addlinespace
\hline
\end{array}
\]
\end{table*}

\begin{table}[htbp]
\centering
\caption{Classification of decision variables using difficulty separation strategy in BOBD}\label{tab4}%
\begin{tabular}{|>{\centering}p{0.8cm}|l|l|}
\hline\multirow{2}{*}{\begin{tabular}{l} 
\textbf{Test} \\
\hspace{-3mm}\textbf{Problem}
\end{tabular}} & \multicolumn{2}{c|}{ \textbf{Variables classified in} } \\
\cline { 2 - 3 } & \multicolumn{1}{c}{ \textbf{Upper Level} } & \multicolumn{1}{|c|}{ \textbf{Lower Level} } \\
\hline
TP1 & $x_1, x_2, x_3$ & $x_4, x_5, x_6$ \\
TP2 & $x_2, x_3, x_4$ & $x_1, x_5, x_6, x_7$ \\
TP3 & $x_1, x_2, x_3$ & $x_4, x_5, x_6$ \\
TP4 & $x_1, x_2, x_3$ & $x_4, x_5, x_6, y_p$ \\
TP5 & $x_2, x_3, x_4$ & $x_1, x_5, x_6, x_7, y_p$ \\
TP6 & $x_3, x_5$ & $x_1, x_2, x_4, x_6, y_p, z_q$ \\
TP7 & $x_8, x_9$ & $x_1, x_2, x_3, x_4, x_5, x_6, x_7, y_p, z_q$ \\
TP8 & $x_1, x_2, x_3, x_4$ & $x_5, x_6, x_7, x_8, x_9, x_{10}, x_{11}, x_{12}, x_{13}, y_p$ \\
TP9 & $x_1, x_2, x_3$ & $x_4, x_5, x_6, x_7, x_8, y_p$ \\
TP10 & $x_1, x_3, x_5$ & $x_2, x_4, y_p, z_q$ \\
\hline
\end{tabular}
\end{table}

\section{Description of Test Problems}
In this study, we use a test suite of 10 test problems to numerically evaluate the proposed ideas. All the test problems in this study are from a real-world context. For the purpose of evaluating our approach we have modified some of these real-world problems by addition further complexities. Among these 10 test problems, there are 2 test problems (TP1-TP2) (\cite{stephanopoulos1975use}-\cite{Liang2006ProblemDA}) that we use directly from the literature. We generate a set of 7 test problems (TP4-TP10) that are scalable in term of variables and constraints. The new test problems (TP3-TP10) are generated by building up on a few optimization problems from the literature \cite{Liang2006ProblemDA}.

Test problems TP3-TP10 have been developed, by considering the general guidelines in developing test problems, such that each of them has one or more complexities, like non-separability, multi-modality, isolation and deception \cite{deb2005scalable,huband2006review,sinha2014test}. These structural properties generate various kinds of difficulties such as non-convexity, discontinuity, discreteness, non-differentiability, which hinder convergence to the optimal solution in the search space. The formulations for the 10 test problems are presented in Tables \ref{tab1}, \ref{tab2} and, \ref{tab3}.

In the test problem set, transcendental functions (\textit{cosine, tangent, logarithmic}) are used to induce difficulties such as non-convexity (TP1-TP10), multi-modality (TP3-TP10), non-separability (TP4-TP10), deception \& isolation (TP8), and discontinuity \& non-differentiability (TP9). In TP4-TP10, terms with variables $y_p$ and/or $z_q$ (where $y=\{y_p : p \in \{1,\ldots,P\}\}$ and $z=\{z_q : q \in \{1,\ldots,Q\}\}$) are introduced to make the test problems scalable in terms of variables and constraints. These test problems are employed to analyze the convergence property of the BOBD method for a large number of variables and constraints (i.e., assessing the scalability of the BOBD method).


\section{Numerical Experiments}
In this section, we compare three approaches: mathematical programming, evolutionary computation, and bilevel optimization-based decomposition. We consider sequential quadratic programming method (SQP) and interior-point method (IP) as mathematical programming approaches \cite{bazaraa2013nonlinear}, and genetic algorithm (GA) from \cite{my-joh20} as the representative evolutionary optimization method. BOBD approach has been implemented within the BLEAQ-II framework \cite{my-joh20}, where we use a classical approach (interior-point) at the lower level and the default GA at the upper level. BLEAQ-II allows us to exploit two important bilevel mappings, i.e., the lower level reaction set mapping and the lower level optimal value function mapping. The choice to use linear programming or non-linear programming approach at the lower level of BOBD is made based on the structure of the lower level problem. The evolutionary algorithm for BOBD and GA is kept the same for a fair comparison. All the methods were encoded in MATLAB and solved in the system with following configuration: Xeon(R) Gold, 64-bit, 32GB RAM, and 160GB storage. In all test problems, for each equality and inequality constraint, the constraint tolerance was set to \(10^{-2}\). Further, for each test problem (TP1-TP10), we execute each algorithm 11 times and report the best feasible solution obtained. 

 While solving any of the test problems using BOBD, each of its decision variables and constraints are classified in the upper or the lower level category. The separation strategy for each test problem is shown in Table \ref{tab4}. For all the test problems (TP1-TP10), after classifying the decision variables into upper and lower categories, all the constraints are part of the lower level; therefore, the classification of decision variables only is presented in Table \ref{tab4}. For the runs of BOBD, we use the population size as 25 times the number of variables at the upper level and terminate the algorithm when there is no improvement in the last 50 generations. For GA, we use a population size of 500 for all test problems and run the approach for the same duration as BOBD. In case of SQP and IP we use the default termination criteria implemented in MATLAB.

 We analyze the result by dividing the test problems into 3 test beds. The first test bed consists of problems TP1-TP10, the second test bed consists of problems TP4-TP10 with dimension more than 20, and the third test bed consists of problem TP4-TP10 with dimension more than 50. Note that we set the total dimensions of scalable variables $y$ and $z$ in TP4-TP7 as 0, 20 and 50, to get the three test beds. The results for each test bed are reported through Tables \ref{tab5}, \ref{tab6}, and \ref{tab7}, respectively.

\begin{table}[htbp]
\centering
\caption{Results for problems with $|y|+|z|=0$}\label{tab5}
\tabcolsep 1mm
\begin{tabular}{|c|l|l|p{1.7cm}|l|l|}
\hline 
\multirow{3}{*}{\begin{tabular}{c} 
\textbf{Test} \\
\textbf{Problem}
\end{tabular}} 
& \multicolumn{4}{c|}{ \textbf{Objective value obtained using} } 
& \multirow{3}{*}{\begin{tabular}{c} 
\textbf{Best Known} \\
\textbf{Solution}
\end{tabular}} \\
\cline{2-5} 
& \multicolumn{2}{c|}{\text{MP} } 
& \multirow{2}{*}{\begin{tabular}{c}
\hspace{4mm}\text{EV} \\
\hspace{4mm}$G A$
\end{tabular}} 
& \multirow{2}{*}{\begin{tabular}{c} 
\text{BO} \\
$B O B D$
\end{tabular}} 
& \\ 
\cline{2-5} 
& $S Q P$ & $I P$ & & & \\
\hline TP1 & -13.40 & -13.40 & -12.401 & -13.417 & -11.96 \\
\hline TP2 & 193.782 & 193.782 & IFL & 189.045 & 193.724 \\
\hline TP3 & -14.401 & -14.401 & -10.896 & -14.401 & -- \\
\hline TP4 & -16.651 & -16.651 & -14.586 & -16.695 & -- \\
\hline TP5 & 143.790 & 143.790 & IFL & 141.178 & -- \\
\hline TP6 & IFL & IFL & -302.676 & -310 & -310 \\
\hline TP7 & -380.32 & -247.93 & IFL & -385.759 & - \\
\hline TP8 & -92.22 & -104.66 & -101.093 & -104.761 & -- \\
\hline TP9 & 7049.2 & 7049.2 & 10345 & 6912.51 & 7049.330 \\
\hline TP10 & -30666 & -30666 & -30663 & -30666 & -30665.538 \\
\hline \multicolumn{6}{|l|}{\fontsize{7}{8}\selectfont MP- Mathematical Programming; EV- Evolutionary; BO- Bilevel Optimization;}\\
\multicolumn{6}{|l|}{\fontsize{7}{8}\selectfont SQP- Sequential Quadratic Programming; IP- Interior Point; GA- Genetic Algorithm;}\\
\multicolumn{6}{|l|}{\fontsize{7}{8}\selectfont BOBD- Bilevel Optimization based Decomposition; IFL- Infeasible.}\\
\cline{1-6}
\end{tabular}
\end{table}


\begin{table}[htbp]
\centering
\caption{Results for problems with $|y|+|z|=20$}\label{tab6}
\begin{tabular}{|c|l|l|p{1.6cm}|l|}
\hline 
\multirow{3}{*}{\begin{tabular}{c} 
\textbf{Test} \\
\textbf{Problem}
\end{tabular}} 
& \multicolumn{4}{c|}{ \textbf{Objective value obtained using} } \\
\cline{2-5} 
& \multicolumn{2}{c|}{\text{MP} } 
& \multirow{2}{*}{\begin{tabular}{c}
\hspace{4mm}\text{EV} \\
\hspace{4mm}$G A$
\end{tabular}} 
& \multirow{2}{*}{\begin{tabular}{c} 
\text{BO} \\
$B O B D$
\end{tabular}} 
\\ 
\cline{2-5} 
& $S Q P$ & $I P$ & & \\
\hline
 TP4 & -10.900 & -10.778 & IFL & -12.201 \\
 TP5 & 189.923 & 189.923 & IFL & 189.854 \\
 TP6 & IFL & -321.60 & -285.348 & -460 \\
 TP7 & -337.91 & -241.44 & IFL & -395.348 \\
 TP8 & -144.66 & 116.44 & -94.952 & -170.346 \\
 TP9 & 7083 & 7064.2 & IFL & 6943.829 \\
 TP10 & 74552 & 74552 & 78877 & 74552 \\
 \hline
\end{tabular}
\end{table}


\begin{table}[htbp]
\centering
\caption{Results for problems with $|y|+|z|=50$}\label{tab7}
\begin{tabular}{|c|l|l|p{1.4cm}|l|}
\hline 
\multirow{3}{*}{\begin{tabular}{c} 
\textbf{Test} \\
\textbf{Problem}
\end{tabular}} 
& \multicolumn{4}{c|}{ \textbf{Objective value obtained using} } \\
\cline{2-5} 
& \multicolumn{2}{c|}{\text{MP} } 
& \multirow{2}{*}{\begin{tabular}{c}
\hspace{2.5mm}\text{EV} \\
\hspace{2.5mm}$G A$
\end{tabular}} 
& \multirow{2}{*}{\begin{tabular}{c} 
\text{BO} \\
$B O B D$
\end{tabular}} 
\\ 
\cline{2-5} 
& $S Q P$ & $I P$ & & \\
\hline
 TP4 & -12.162 & -12.151 & IFL & -12.133 \\
 TP5 & 259.125 & 264.585 & IFL & 259.125 \\
 TP6 & IFL & -541.084 & -253.246 & -685 \\
 TP7 & -300 & -332 & IFL & -404.825 \\
 TP8 & -268.792 & -239.765 & 38.735 & -289.488 \\
 TP9 & 7816.9 & 8176.8 & IFL & 7012.415 \\
 TP10 & 232380 & 232380 & 245360 & 232380 \\
 \hline
\end{tabular}
\end{table}

In Table \ref{tab5} ($|y|+|z|=0$), the best known solutions of TP1, TP2, TP6, TP9, and TP10 are provided. Note that $|.|$ denotes the dimensionality of the decision variables. For TP1, all the methods have provided improved solution compared to its best known solution. In the case of both TP1 and TP2, the solution from BOBD is better than other methods and also better than the solutions reported in the literature. The solutions provided by BOBD for other test problems are either equal or better compared to the best known solution of the respective problems. In none of the cases BOBD is outperformed by SQP, IP or GA in terms of solution quality. The methods from mathematical programming approach, SQP and IP, succeed in providing solutions close to the best known solution for most of the test problems, except for TP6, where both methods delivered infeasible solutions. GA delivered infeasible solutions for TP2, TP5, and TP7, and for the other test problems, the obtained solutions are worse compared to other methods. 

Based on results in Table \ref{tab6} ($|y|+|z|=20$), which contains scaled test problem, it is observed that BOBD outperforms the other methods by a considerable margin in the case of all test problems (TP4-TP10). The same observation is valid for Table \ref{tab7} ($|y|+|z|=50$). For all the runs, in our study, BOBD always delivered feasible solutions for all the test problems in each test bed, whereas other methods provided infeasible (IFL) solutions for a number of runs. The best, median and the worst results of BOBD also coincided for the 11 runs, which emphasises the consistency of the approach. 
Using an evolutionary approach (GA) alone provided infeasible solutions more times compared to other methods, which highlights the commonly known deterioration in performance of EAs when the dimensions are high or there are multiple equality or inequality constraints. 


\section{Conclusions}
This study introduces an approach for tackling single level complex optimization problems through a Bilevel Optimization based Decomposition (BOBD) method. The proposed strategy systematically categorizes each decision variable and constraint of a given optimization problem into upper or lower level, thereby reformulating the single level optimization problem into a bilevel optimization problem. This transformation allows the utilization of two suitable approaches to synergistically solve the optimization problem. In this study, we solve a number of challenging optimization problem using BOBD, which are otherwise difficult to solve using a classical optimization or evolutionary optimization approach alone.

The proposed approach is applied to a test suit of 10 test problems, where a few problems are taken from the literature, and some have been extended as a part of this study to introduce various kinds of difficulties. The performance of BOBD is benchmarked against mathematical programming-based approaches, interior point and sequential quadratic programming methods, as well as an evolutionary approach. Comparative analyses across the 10 test problems with various dimensions reveal that BOBD consistently outperforms the other methods in terms of solution quality. As a part of our future study, we aim to do the classification of variables and constraints into upper and lower levels automatically and dynamically, in lieu of manually prescribing them.

\section*{Acknowledgement}
The last author acknowledges support from Discovery Project DP220101649 from the Australian Research Council.

\balance
\bibliographystyle{ieeetr}
\bibliography{references}

\end{document}